\documentclass[10pt,twocolumn,letterpaper]{article}

\newif\ifarxiv
\arxivfalse

\newif\ifreview
\reviewfalse

\ifarxiv
\usepackage[pagenumbers]{cvpr} 
\else
\ifreview
\usepackage[review]{cvpr}      
\else
\usepackage{cvpr}              
\fi
\fi    

\usepackage{multirow}
\usepackage{array}
\usepackage{siunitx}
\usepackage[table]{xcolor} 
\usepackage{hhline}
\usepackage{pifont}
\usepackage{bbding}
\usepackage{balance}

\definecolor{cvprblue}{rgb}{0.21,0.49,0.74}
\usepackage[pagebackref,breaklinks,colorlinks,allcolors=cvprblue]{hyperref}

\def\paperID{***} 
\def\confName{CVPR}
\def\confYear{2026}

\ifreview
\title{Unified Gaze Estimation Framework for Mixed Reality with a Large-Scale Multi-View Dataset}
\else
\title{PicoEyes: Unified Gaze Estimation Framework for Mixed Reality with a Large-Scale Multi-View Dataset}
\fi

\ifarxiv
\author{Preprint}
\else
\author{Fuxin Duan \quad Hui Wang \\
Pico, Bytedance\\
{\tt\small \{duanfuxin, wanghui2207\}@bytedance.com}\\
}
\fi

\begin{document}
\maketitle
\begin{abstract}

\ifreview
\else
We present PicoEyes, a unified gaze estimation framework
\fi
that directly predicts all key attributes of gaze, including 3D eye parameters, eye-region segmentation, optical axis, visual axis, and depth maps, from either monocular or binocular inputs. The framework simultaneously addresses calibration, gaze forecasting, and varying device postures, while also supporting 3D eye reconstruction via joint estimation of eye parameters and depth maps in an end-to-end manner. In addition, we introduce a large-scale multi-view near-eye dataset containing comprehensive 2D and 3D annotations under diverse conditions, including train, test, rewear-test, and calibration sessions. 
\ifreview
\else
Extensive experiments demonstrate that PicoEyes achieves state-of-the-art performance,
\fi
consistently outperforming both academic and industrial gaze tracking methods across no-calibration, calibration, rewear-after-calibration, and forecasting settings. This work establishes a practical, end-to-end paradigm for robust and generalizable gaze estimation in mixed reality (MR) applications.

\end{abstract}

\section{Introduction}
\label{sec:intro}

Eye tracking, which seeks to estimate human visual attention, has emerged as a key technology for MR applications such as hand–eye interaction \cite{pfeuffer2017gaze+}, eye-tracked foveated rendering \cite{liu2025fovealnet}, auto-focus \cite{chakravarthula2018focusar}, and ET-LLM/VLM \cite{rekimoto2025gazellm, yan2024voila}.

Classic approaches can be broadly categorized into model based and appearance based methods. Model-based \cite{guestrin2006general} methods rely on explicit 3D eye geometry structure and physical constraints, providing potentially high precision but often requiring specialized hardware. In contrast, appearance-based \cite{krafka2016eye, zhang2020eth, palmero2020openeds2020, kim2019nvgaze, kellnhofer2019gaze360, zhang2017s, wang2024pveye} methods generally utilize deep neural networks to directly estimate gaze from eye or face images, providing greater scalability and flexibility without compromising precision.

\begin{figure}
   \includegraphics[width=1.0\linewidth]{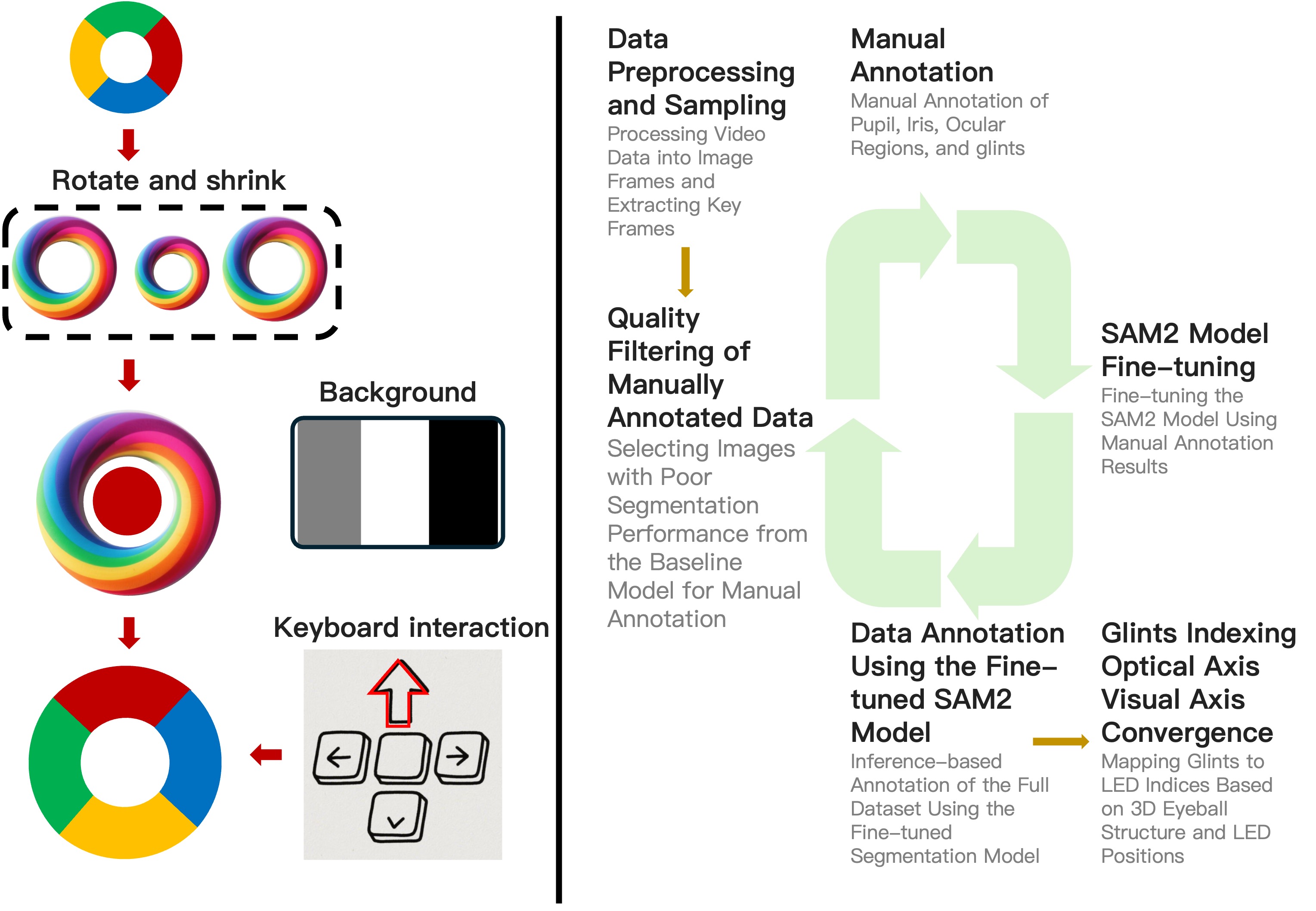}
   \caption{Virtual target animation/interaction (left) and data annotation pipeline (right).}
   \label{fig:recording_and_target}
\end{figure}

Despite the release of numerous open-source MR et datasets (\eg, NVGaze \cite{kim2019nvgaze}, OpenEDS \cite{palmero2020openeds2020}, TEyeD \cite{fuhl2021teyed}, MagicEyes \cite{wu2020magiceyes}, PVEye \cite{wang2024pveye}), existing datasets still present notable limitations. Most are split into training and testing subsets with either gaze or segmentation labels or occasionally both, yet annotations of explicit eyeball parameters are almost entirely absent. Furthermore, practical issues such as user-specific calibration and performance degradation after re-wearing a head-mounted device remain largely unaddressed. In immersive virtual reality(VR) scenarios, cameras are often embedded within pancake optics, making camera parameterization challenging or infeasible, similar problems occur when users wear glasses. Finally, current datasets and methods rarely support multi-view gaze estimation, which is essential for many MR applications.

Furthermore, deep learning-based gaze estimation methods typically directly regressing gaze from eye images, predicting eye-region segmentation, or using separate models for specific subtasks. In multi-view settings, multiple views are exploited via stereo rectification \cite{cheng2023dvgaze} or feature fusion through extrinsic transformations \cite{hisadome2024rotation}. The former is impractical when camera parameters cannot be explicitly specified, as in VR systems with pancake optics, the latter focuses on extrinsic-driven alignment and underutilizes complementary cross view information. So, there remains a lack of an end-to-end unified paradigm that can jointly address all task in gaze estimation,  adapt to both single-view and multi-view scenarios, and are compatible with user calibration alongside gaze forecasting. In addition, compared with model-based approaches, current methods deliver lower accuracy($\geq 2\si{\degree}$), which is insufficient to enable high-precision gaze-driven interaction in MR applications.

To address these challenges, we propose a unified, end-to-end gaze estimation framework that can seamlessly process both binocular and monocular inputs. The framework jointly reconstructs the full 3D eye structure, performs gaze calibration, and conducts gaze forecasting, leveraging multi-view feature fusion via an epipolar constraint attention. We further present a large-scale multi-view eye dataset comprising data from 617 participants, annotated with exhaustive 2D and 3D labels, and covering diverse conditions including train, test, rewear test, and calibration, enabling more rigorous and comprehensive model development.

Extensive experiments demonstrate that \textbf{our approach achieves state-of-the-art performance in gaze tracking. Moreover, evaluations on our proprietary datasets show that the proposed method consistently achieves higher accuracy across no-calibration, calibration, rewear after calibration, and forecasting settings, making it suitable for real-time deployment in MR devices.} In summary, our main contributions are as follows:

\begin{itemize}
    \item We release a large-scale multi-view eye dataset with comprehensive 2D and 3D annotations, enabling robust model training and evaluation under diverse conditions.
    \item We propose a unified end-to-end gaze estimation framework that jointly performs gaze prediction, eye segmentation, gaze calibration, and gaze forecasting, while supporting both binocular and monocular inputs.
    \item We achieve state-of-the-art performance on proprietary datasets, with demonstrated applicability to MR.
\end{itemize}
\section{Related Work}
\label{sec:relatedwork}



\subsection{Gaze Datasets}

Due to the difficulty of collecting high precision near-eye gaze data, publicly available datasets are scarce and often lack complete eye information, limiting their effectiveness for model training. Existing datasets such as NvGaze \cite{kim2019nvgaze}, OpenEDS \cite{garbin2019openeds, palmero2020openeds2020}, MagicEyes \cite{wu2020magiceyes}, and others provide synthetic or real-world data but include only gaze and pupil localization or segmentation annotations, lacking more comprehensive details. TEyeD \cite{fuhl2021teyed} offers both 2D and 3D annotations but covers only 132 subjects, while PVEye \cite{wang2024pveye} additionally considers the effects of eyewear. Except for MagicEyes, these datasets involve few subjects and overlook critical factors such as test subject diversity, calibration, rewear after calibration, and time series. Our dataset addresses these limitations by incorporating a larger and more diverse participant alongside complete, temporally eye data under varied conditions. \cref{tab:dataset} provides a comprehensive comparison between our dataset and existing ones.

\subsection{Gaze Estimation Algorithms}

Gaze estimation methods can broadly be categorized into model-based and appearance-based algorithm. Model-based \cite{guestrin2006general, wood20163d, hansen2009eye} gaze estimation leverages geometric eye models and physical constraints to attain high accuracy, and its dependence on hardware configurations (\eg, infrared light and camera) restricts use in open environment for MR and limits generalization. In contrast, appearance-based methods \cite{zhang2015appearance, zhang2020eth, cheng2024appearance} learn a mapping from image features to gaze via machine learning, offering greater flexibility and significant potential for generalization. These strategies generally fall into three main categories:
\textbf{(1) Direct regression} from raw images to gaze vectors \cite{kim2019nvgaze, linden2019learning, palmero2020openeds2020, wu2020magiceyes, wang2024pveye},
\textbf{(2) Hybrid geometric–learning}, such as eye segmentation, glint or pupil localization, from which gaze is subsequently derived using purely geometric modeling \cite{wu2019eyenet}, and
\textbf{(3) Eye parameter regression} (\eg, eyeball center and optical axis), which are then transformed into visual axes via calibration for the kappa angle \cite{fuhl2021teyed, bao2025gazegene}. In summary, despite substantial progress, a unified framework capable of robustly handling all gaze estimation challenges remains elusive, especially for calibration.

\ifreview
\else
\section{PicoEyes Dataset}
\fi
\label{sec:picoeyesdatasets}

\begin{table*}[]
\caption{Comparison of eye-tracking datasets for MR devices. Our dataset provides more comprehensive information required for practical MR eye-tracking.}
\label{tab:dataset}
\centering
\renewcommand{\arraystretch}{1.1}
\begin{tabular*}{\textwidth}{@{\extracolsep{\fill}}lcccccccccc}
\specialrule{1.2pt}{0pt}{0pt}
\multirow{3}{*}{Data} & \multirow{3}{*}{Sub.} & \multirow{3}{*}{Total} & \multirow{3}{*}{Res.} & \multirow{3}{*}{FRQ} & \multicolumn{6}{c}{Annotation} \\ 
\cline{6-11} 
                      &                       &                        &                       &                      & \begin{tabular}[c]{@{}c@{}}Gaze Range\\ (yaw, pitch)\end{tabular} & Vergence & \begin{tabular}[c]{@{}c@{}}Eye\\ Seg.\end{tabular} & \begin{tabular}[c]{@{}c@{}}3D \\ Eye\end{tabular} & \begin{tabular}[c]{@{}c@{}}Led\\ Loc.\end{tabular} & \begin{tabular}[c]{@{}c@{}}Cam \\ Param.\end{tabular} \\ 
\specialrule{0.8pt}{0pt}{0pt}
NVGaze \cite{kim2019nvgaze}                & 35   & 2.5M  & 480$\times$640    & 30Hz  & 40°, 30°   & ------   & \color{red}\ding{55}   & \color{red}\ding{55}  & \color{red}\ding{55} & \color{red}\ding{55}   \\
OpenEDS \cite{garbin2019openeds}           & 152  & 3.5M  & 400$\times$640    & 200Hz & ------     & ------   & \color{green}\checkmark  & \color{green}\checkmark & \color{red}\ding{55} & \color{red}\ding{55}   \\
OpenEDS2020 \cite{palmero2020openeds2020}  & 80   & 5.8M  & 400$\times$640    & 100Hz & ±20°, ±20° & 0.5-6m   & \color{green}\checkmark  & \color{red}\ding{55}  & \color{red}\ding{55} & \color{red}\ding{55}    \\
TEyeD \cite{fuhl2021teyed}                 & 132  & 20.8M & 480$\times$640    & 120Hz & ------     & ------   & \color{green}\checkmark  & \color{green}\checkmark & \color{red}\ding{55} & \color{red}\ding{55}   \\
MagicEyes \cite{wu2020magiceyes}           & 587  & 80M   & 480$\times$640    & 30Hz  & ------     & 0.5-3m   & \color{green}\checkmark  & \color{green}\checkmark & \color{red}\ding{55} & \color{red}\ding{55}   \\
PVEye \cite{wang2024pveye}                 & 104  & 11M   & 400$\times$640    & 30Hz  & ±20°, ±12° & 3.1m     & \color{red}\ding{55}   & \color{red}\ding{55}  & \color{red}\ding{55} & \color{red}\ding{55}   \\
\ifreview
\else
\textbf{Picoeyes}                         & \textbf{617} & \textbf{2.5M}      & \textbf{480$\times$640}      & \textbf{90Hz}  & \textbf{±30°, ±30°} & \textbf{0.9-1.5m}  & \color{green}\checkmark  & \color{green}\checkmark & \color{green}\checkmark & \color{green}\checkmark \\
\fi
\specialrule{1.2pt}{0pt}{0pt}
\end{tabular*}
\end{table*}

\subsection{Platform and Procedure}

Choosing a suitable hardware data platform is crucial for gaze estimation dataset acquisition. We selected the custom built head-mounted MR device from Pico as our hardware acquisition platform. The MR headset contains two optical barrels, each equipped with two high‑resolution infrared eye tracking cameras. The camera offers an 80° diagonal field of view, 640$\times$480 resolution, and runs at 90Hz. Each optical barrel is equipped with 14 LEDs that provide controlled eye illumination and produce cornea reflection patterns. Together with the dataset, we provide precise camera intrinsic and extrinsic parameters, as well as the coordinates of the LEDs.

\noindent \textbf{Data Collection Platform.} To evaluate various aspects of the gaze estimation algorithm, we developed a data acquisition tool in Unity. This tool is capable of presenting targets at different positions and depths on a virtual space within the MR. Participants were instructed to fixate on each target and perform corresponding action according to predefined requirements. In addition, the tool can simulate different lighting conditions, ensuring that the pupillary responses of the participants remain consistent with those of the real‑world usage scenarios. 


\noindent \textbf{Visual Targets.} Participants were required to fixate on visual targets. To ensure fixation during data collection, each target incorporated specific visual cues(\cref{fig:recording_and_target}). A circular ring surrounding the target was evenly divided into four colors and, at the start of acquisition, rotated and blinked to attract the participants’ attention. A random selected color then appeared at the center, the ring halted in a random orientation, and the central color disappeared. Participants should use keyboard arrow keys or a controller joystick to indicate the matching color position on the ring. If the participant matched the correct color location in all trials, it was deemed that the participant was fixating on the target and this point was considered as a valid sample. Due to the sensitivity of pupil size to illumination, as shown in \cref{fig:recording_and_target}, we designed three background colors(black, gray, and white), to simulate realistic usage scenarios, each corresponding to a different pupil size.

\noindent \textbf{Gaze Range.} Head pose and gaze range are fundamental attributes of a gaze dataset. In this study, we use an MR device, and gaze direction is reported in the device coordinate system. Within this system, visual targets were randomly positioned within a horizontal and vertical range of [-30$\si{\degree}$, 30$\si{\degree}$]. To simulate eye states at varying viewing distances, visual targets were also randomly distributed within a depth range of 900$\sim$1500mm. The target size was proportional with viewing distance, yielding a theoretical gaze accuracy of approximately 0.15$\si{\degree}$.


\noindent \textbf{Participants.} A total of 617 participants were recruited, including 260 males and 357 females, aged between 18 and 50 years. Among them, 247 participants had non-myopia, 370 participants had myopia ranging from 0 to 950 degrees. All participants provided written informed consent prior to data collection, agreeing to the use of their eye images for both scientific research and commercial purposes. 

\noindent \textbf{Data Collection Procedure.} Each participant completed data collection using six devices, with six groups for per device. Participants were assigned randomly to either train or test set, ensuring that no individual appeared in both. For the train set, the head-mounted configuration was adjusted after each group (\eg, rewear or change nose pad). For the test set, calibration and test were alternated, after completing each calibration–test pair, and the head-mounted configuration was adjusted too.


\subsection{Dataset annotation}

For annotations, we provide high quality pixel level mask for sclera, iris, pupil and glint. In addition, we also offer accurate 3D annotations of the eye structures.

\noindent \textbf{Eye Segmentation.} The collected eye video data are large in volume and contain substantial redundancy, with numerous fixation states producing frames that are nearly static or exhibit regular patterns. Manual annotation of such data is time-consuming and labor-intensive, making it impractical for real-world applications.

For segmentation annotation, we employ the Segment Anything Model 2 (SAM2) \cite{ravi2024sam} to perform preliminary segmentation on sampled data, thereby reducing labeling costs and improving efficiency. Images with poor segmentation quality are subsequently classified and manually annotated. During manual annotation, we focus on key frames where the baseline SAM2 model \cite{ravi2024sam} performs poorly, ensuring complete labeling of the sclera, pupil, iris, and glint. These manually annotated key frames are then used to finetune the SAM2 model, which is ultimately applied to annotate the entire video dataset.

We evaluated the finetuned SAM2 model on a subset of manually annotated data that was excluded from training. Compared to the manual annotations, the finetuned model achieved an Intersection over Union (IoU) of 0.89. Qualitative segmentation results are presented in Appendix.



\noindent \textbf{Eye Parameters and Glints Match.} Based on the 3D anatomy of the eye, we approximate its structure using a simplified geometric model in which a small spherical cornea rotates about a larger spherical eyeball. When illuminated by LED point light sources, several reflection glints are formed on the corneal surface. The imaging of these glints follows the principles of mirror reflection, with the resulting reflections jointly determined by the positions of the light sources, the camera, and the cornea.

Given prior information on the relative positions of the LEDs and the camera, together with the observed locations of corneal glints in the captured images, variations in the corneal position yield different imaging outcomes. We estimate the corneal center, corneal radius, and the mapping between glints and their corresponding LEDs by iteratively searching for a spatial configuration whose simulated imaging results best match the observed data. Through multi‑frame joint optimization, the eyeball center and radius can also be determined.

\noindent \textbf{Gaze and Convergence.} Based on the annotated glints and eye geometry, the 3D positions of the corneal center and the eyeball center are estimated for each frame. The optical axis is defined as the vector extending from the eyeball center to the corneal center and the visual axis is defined as the vector from the corneal center to the virtual target position. Convergence is computed as the Euclidean distance between the average location of the left and right corneal centers and the virtual target.

\ifreview
\else
\section{PicoEyes}
\fi
\label{sec:picoeyes}

\begin{figure*}[t]
  \centering
   \includegraphics[width=1.0\linewidth]{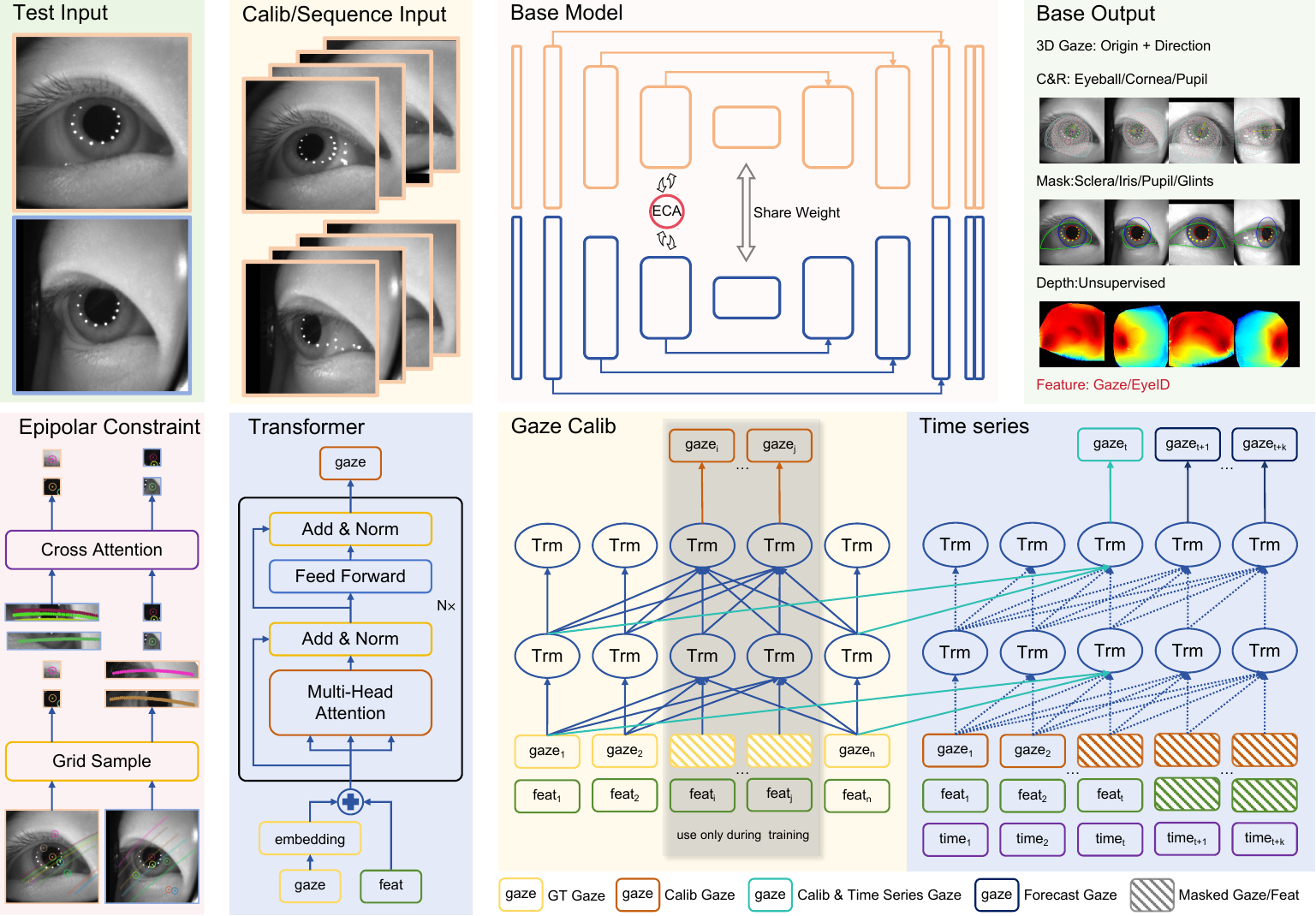}
   \ifreview
\else
    \caption{\textbf{Overall architecture of the proposed \textit{PicoEyes} framework.} (a) The current input is processed by the base model (b) to estimate (d) \emph{uncalibrated} gaze and gaze origin, and to produce auxiliary outputs (eye parameters, segmentation, depth) as well as features for calibration and temporal module. For calibration/temporal reference inputs, the base model extracts and caches features, which are then fed into (g) GazeCalib and (h) TimeSeries to output calibrated gaze and future gaze predictions, respectively. (e) The ECA module fuses features when multi-view eyes are available. In (g) GazeCalib, gaze inputs are randomly masked during training, while only the last gaze is masked at inference. (h) TimeSeries predicts gaze for the next $k$ time steps.}
   \fi
   \label{fig:picoeyes}
\end{figure*}

\ifreview
\else
We introduce \textit{PicoEyes}, a unified gaze estimation paradigm capable of processing both binocular and monocular eye inputs in an end-to-end fashion.
\fi
The system accepts a set of eye images as input and produces diverse 2D/3D gaze related quantities as output. An overview of the proposed framework is illustrated in \cref{fig:picoeyes}. 

\subsection{Problem Definition}
\label{sec:problemdefinition}


The inputs consist of a test dataset and multiple calibration/temporal datasets. Each dataset contains a pair of images captured from the left and right eyes, acquired from either a single view or multiple views, and denoted as $I^t=\{I_c^{t}\}_{t\in \{l, r\}, c\in[1,M]}$, where $t\in \{l,r\}$ indicates the left or right eye respectively, $I_c^t\in\mathbb{R}^{H\times W}$ is the image captured from view \(c\), and \(M\) is the total number of available views. In addition, each set is associated with the corresponding camera intrinsic and extrinsic parameters $C^t = \{C_c\}_{c=1}^{M}$.

\ifreview
\else
\textit{PicoEyes} is a mapping function that, for each inference, takes as input the test images and the calibration sequences, together with the camera parameters, and produces both per-eye and per-view outputs. 
\fi

\begin{equation}
  f(I^t, (I^t_i)_{i=1}^N, C^t) = (O^t, G_o^t, G_v^t, E^t, S^c, D^c).
  \label{eq:map_function}
\end{equation}

In our formulation, for per-eye outputs, the \textbf{gaze} is parameterized by its origin and direction, \ie, expressed as a 3D point $O^t\in \mathbb{R}^3$ indicating the starting position(corneal center) and a normalized vector specifying the gaze direction $G^t\in \mathbb{R}^3$. Importantly, like in \cite{guestrin2006general}, the gaze directions $G_{o}$ and $G_{v}$ correspond to the optical axis and the visual axis of the eye, respectively. As for \textbf{eye parameters} $E \in \mathbb{R}^{12}$, we use the parametrization from \cite{bao2025gazegene} which is the concatenation of the eyeball center $p_e \in \mathbb{R}^3$ and radius $r_e \in \mathbb{R}$, the corneal center $p_c \in \mathbb{R}^3$ and radius $r_c \in \mathbb{R}$, and pupil center $p_p \in \mathbb{R}^3$ and radius $r_p \in \mathbb{R}$.

For the per-view outputs, the \textbf{semantic segmentation} $S^c$ for view $c$, includes sclera, cornea, pupil, and infrared reflection glints. The \textbf{depth map} \(D^c\) provides dense 3D geometric information that complements the reconstruction of eye region, enabling 3D digital human eye modeling.

\subsection{Feature Backbone}
\label{sec:featurebackbone}

Following recent works in appearance-based gaze estimation using deep learning \cite{zhang2015appearance, zhang2017mpiigaze, guo2019generalized, zhang2020eth, hisadome2024rotation, bao2025gazegene}, we adopt a CNN backbone architecture and employ a UNet decoder for dense prediction tasks as shown in \cref{fig:picoeyes} first row.

\noindent\textbf{Pixel-Wise Camera Parameters}. Since the camera is installed inside the lens barrel of the HMD, it introduces distortions and intrinsic parameters that cannot be accurately modeled using conventional parameterization methods. Therefore, we adopt a more general representation(non-central generic model \cite{schops2020having}) of the camera intrinsic parameters to accommodate such complex optical effects. Moreover, Plücker coordinate \cite{jia2020plucker}, a general and elegant formulation for representing camera parameters, is widely used in camera control diffusion model \cite{xu2024camco, he2025cameractrl}. Given intrinsics and extrinsics, camera embedding for each 2d pixel $(u, v)$ is defined as $C(u,v)=\{\mathbf{R}[p(u,v)\times d(u,v)] + \mathbf{T}, \mathbf{R}d(u,v))\} \in \mathbb{R}^6$, and all camera poses are defined relatively to the device coordinates.

\noindent\textbf{Epipolar-Constraint Attention}. For multi-view, due to the unique characteristics of the camera parameters in our system, the epipolar lines become spline curves rather than straight lines(\cref{fig:picoeyes}), resulting conventional stereo rectification methods \cite{guo2024stereo} inapplicable. Furthermore, multi view stereo solutions, such as MVSNet \cite{yao2018mvsnet}, are infeasible in our system due to their prohibitive computational cost.


We adopt the Epipolar-Constraint Attention(ECA) proposed in \cite{xu2024camco}. At each CNN stage, we represent $z \in \mathbb{R}^{hw \times d}$ as the feature map, and $Z \in \mathbb{R}^{hw\times l \times d}$ as the point feature along the corresponding epipolar lines, where $l$ is the number of points on the epipolar lines, extracted from the feature map of another view. The query, key, value are given by $q=zW_q\in \mathbb{R}^{hw\times 1 \times d}$, $k=ZW_k\in \mathbb{R}^{hw\times l \times d}$, $v=ZW_v\in \mathbb{R}^{hw\times l \times d}$. The ECA between two frame is then computed as $\mathrm{ECA}(z, Z) = \sigma(\frac{qk^T}{\sqrt{d}})v \in \mathbb{R}^{hw \times d}$.

\subsection{Prediction Heads}
\label{sec:predictionheads}

In this section, we describe the design of the prediction heads. For multi-view inputs, we first apply a model with sharing weights to extract features from each view. These features are then refined using ECA. The multi-view features are fused through concat, followed by a convolutional layer, and finally used to produce the prediction outputs.

\noindent\textbf{Gaze Predictions}. As described above, we directly predict origin point $O^t\in \mathbb{R}^3$, gaze(optical axis) $G_o^t\in \mathbb{R}^3$ and eye parameters $E^{t} \in \mathbb{R}^{12}$ of each eye. The origin loss is defined as  $\mathcal{L}_{\text{origin}}=\lambda_{\text{origin}} \sum_{t\in \{l,r\}}|| O^t - \hat{O}^t ||_2^2$ which supervises the origin points against the ground truth. The optical gaze loss is given by $\mathcal{L}_{\text{optical}}=\lambda_{\text{optical}} \sum_{t\in \{l,r\}}|| G_o^t - \hat{G}_o^t ||_2^2$ and before calculating gaze loss, each gaze vector is converted into pitch and yaw angles. Finally, the eye parameter loss is defined as $\mathcal{L}_{\text{eye}}=\lambda_{\text{eye}} \sum_{t\in \{l,r\}}|| E^t - \hat{E}^t ||_2^2$.

\noindent\textbf{Dense Predictions}. To address the dense prediction tasks of segmentation and depth estimation, we adopt the UNet architecture \cite{ronneberger2015u}, where an encoder–decoder framework is designed to capture multi-scale feature while preserving spatial details through skip connections. 

 \noindent \textbf{Semantic Segmentation}. We employ a combination of Focal Loss \cite{lin2017focal} and Dice Loss \cite{milletari2016v} to address issues such as class imbalance. The overall segmentation loss is formulated as $\mathcal{L}_{\text{seg}} = \lambda_{\text{focal}} \mathcal{L}_{\text{focal}} + \lambda_{\text{dice}} \mathcal{L}_{\text{dice}} \label{eq:seg_loss}$.

For depth estimation, we adopt an unsupervised training strategy that combines multiple complementary loss functions. Specifically, we integrate the Chamfer Distance (CD) \cite{fan2017point}, Normalized Cross-Correlation (NCC) \cite{lewis1995fast}, Sum of Squared Differences (SSD) \cite{horn1981determining}, and the Structural Similarity Index Measure (SSIM) \cite{wang2004image}. This combination enables the model to simultaneously capture geometric consistency, photometric consistency, and perceptual similarity. The overall depth estimation loss is defined as
$\mathcal{L}_{\text{depth}} = \lambda_{\text{cd}}\mathcal{L}_{\text{cd}} + \lambda_{\text{ncc}}\mathcal{L}_{\text{ncc}} + \lambda_{\text{ssd}}\mathcal{L}_{\text{ssd}} + \lambda_{\text{ssim}}\mathcal{L}_{\text{ssim}}$.

\begin{table*}
\caption{\textbf{Gaze estimation under different camera setups, targets, and calibration conditions.} We report \emph{Accuracy}/\emph{Precision}/\emph{Origin}/\emph{Convergence} as (\si{\degree}/\si{\degree}/mm/$\mathrm{d}$). \textbf{Accuracy} is the mean angular gaze error (\si{\degree}); \textbf{Precision} is the variance (\si{\degree}); \textbf{Origin} is the 3D gaze-origin error (mm); \textbf{Convergence} is the vergence error $\lvert D^{-1}-\hat{D}^{-1}\rvert$ ($\mathrm{d}$, diopters). \textbf{Cam.} denotes the camera setup (\textbf{Bino.}/\textbf{Mono.}), and \textbf{T.} denotes the target/eye setting (\textbf{L}: left; \textbf{R}: right; \textbf{C}: combined left and right). We evaluate three modes: \textbf{Calibration}, \textbf{Calibration (Rewear)} (after removing and re-wearing the device), and \textbf{No Calibration}. Each cell shows \textbf{Avg} (top) and \textbf{P90} (bottom, worst-case). Unless noted, all later results use \textbf{Calibration (Rewear)}.}
\label{tab:gaze_results}
\centering
\renewcommand{\arraystretch}{1.1}
\setlength{\tabcolsep}{5pt}
\begin{tabular*}{\textwidth}{@{\extracolsep{\fill}}ccccc}
\specialrule{1.2pt}{0pt}{0pt}
Camera & Tube & Calibration & Calibration(Rewear) & No Calibration \\
\specialrule{0.8pt}{0pt}{0pt}

\multirow{6.5}{*}{Binocular} & \multirow{2}{*}{Combine}   
& \cellcolor{gray!20}\textbf{Avg: 0.42 / 0.23 / 0.12 / 0.23} & \cellcolor{gray!20}\textbf{Avg: 0.44 / 0.22 / 0.12 / 0.21} & \cellcolor{gray!20}\textbf{Avg: 1.25 / 0.28 / 0.12 / 2.29}  \\
&     & P90: 0.54 / 0.27 / 0.20 / 0.18 & P90: 0.56 / 0.27 / 0.20 / 0.18 & P90: 2.30 / 0.35 / 0.20 / 2.96 \\
\cmidrule(r){2-5}

& \multirow{2}{*}{Left}  
& Avg: 0.51 / 0.27 / 0.14 / ------ & Avg: 0.53 / 0.27 / 0.14 / ------ & Avg: 1.77 / 0.36 / 0.14 / ------  \\
&     & P90: 0.68 / 0.33 / 0.22 / ------ & P90: 0.73 / 0.33 / 0.22 / ------ & P90: 2.85 / 0.44 / 0.22 / ------ \\
\cmidrule(r){2-5}

& \multirow{2}{*}{Right}  
& Avg: 0.55 / 0.27 / 0.14 / ------ & Avg: 0.57 / 0.27 / 0.14 / ------ & Avg: 1.80 / 0.36 / 0.14 / ------  \\
&     & P90: 0.69 / 0.34 / 0.24 / ------ & P90: 0.76 / 0.34 / 0.24 / ------ & P90: 2.79 / 0.45 / 0.24 / ------ \\
\specialrule{0.8pt}{0pt}{0pt}

\multirow{6.5}{*}{Monocular} & \multirow{2}{*}{Combine}   
& \cellcolor{gray!20}\textbf{Avg: 0.53 / 0.27 / 0.33 / 0.43} & \cellcolor{gray!20}\textbf{Avg: 0.55 / 0.26 / 0.33 / 0.36} & \cellcolor{gray!20}\textbf{Avg: 1.39 / 0.33 / 0.33 / 3.21}  \\
&     & P90: 0.70 / 0.33 / 0.56 / 0.27 & P90: 0.74 / 0.35 / 0.56 / 0.29 & P90: 2.30 / 0.41 / 0.56 / 4.21 \\
\cmidrule(r){2-5}

& \multirow{2}{*}{Left}  
& Avg: 0.66 / 0.33 / 0.40 / ------ & Avg: 0.69 / 0.33 / 0.40 / ------ & Avg: 1.94 / 0.42 / 0.40 / ------  \\
&     & P90: 0.88 / 0.42 / 0.64 / ------ & P90: 0.93 / 0.42 / 0.64 / ------ & P90: 3.03 / 0.52 / 0.64 / ------ \\
\cmidrule(r){2-5}

& \multirow{2}{*}{Right}  
& Avg: 0.70 / 0.33 / 0.40 / ------ & Avg: 0.73 / 0.33 / 0.40 / ------ & Avg: 1.98 / 0.43 / 0.40 / ------  \\
&     & P90: 0.89 / 0.38 / 0.67 / ------ & P90: 0.95 / 0.42 / 0.67 / ------ & P90: 2.99 / 0.53 / 0.67 / ------ \\
\specialrule{1.2pt}{0pt}{0pt}
\end{tabular*}
{\footnotesize\raggedright *The convergence error is computed as $|D^{-1}-\hat{D}^{-1}|$. Unless otherwise stated, all metrics presented in the remainder of this paper are measured under the rewearing condition.\par}
\end{table*}

\subsection{Gaze Calibration and Forecasting}
\label{sec:gazecalibrationandforecasting}

Considering individual physiological variations in eyeball structure (\eg, the kappa angle \cite{guestrin2006general}), we propose a gaze calibration module (\cref{fig:picoeyes}) that adaptively compensates for systematic biases caused by such physiological differences, thereby enabling accurate inference of the gaze (visual axis). By learning a personalized transformation from captured eye features to the calibrated gaze, our approach effectively reduces estimation errors arising from physiological diversity across users. To the best of our knowledge, we are the first to introduce a deep learning-based, end-to-end gaze calibration framework that eliminates the need for traditional, model-based manual calibration procedures.

\noindent \textbf{Gaze Calibration}. Inspired by the masked autoencoding paradigm exemplified by MAE \cite{he2022masked} and BERT \cite{devlin2019bert}, our gaze calibration method is implemented by leveraging the feature maps produced by a base gaze estimation model. During inference, we take as input the feature map of a test sample, together with the gaze labels and feature maps of $N$ calibration samples. These inputs are fused to estimate the gaze (visual axis) for the test sample. In the training phase, we follow a self-supervised strategy by randomly masking a certain proportion of gaze labels. This design enables the network to learn inter-sample relationships and calibration mappings. The model is then optimized to reconstruct the masked labels from the remaining visible labels and associated feature maps. The visual gaze loss is given by $\mathcal{L}_{\text{visual}}=\lambda_{\text{visual}} \sum_{t\in \{l,r\}}|| G_v^t - \hat{G}_v^t ||_2^2$.


\noindent \textbf{Gaze Forecasting}. We further explore the task of gaze forecasting, where the objective is to anticipate future gaze based on historical gaze trajectories and visual context. This capability reduces the latency in acquiring gaze information and enhances overall estimation accuracy. Motivated by the causal autoregressive modeling paradigm popularized by GPT \cite{radford2018improving}, our framework uses a similar strategy for gaze prediction \cite{jindal2024spatio}. Specifically, the forecasting module generates gaze estimates for the subsequent $K$  discrete time steps, conditioned on past gaze sequences. During inference, computation latency is inevitable. To mitigate this, we obtain the gaze for the intended timestamp by performing temporal interpolation over the $K$ predicted gaze, thereby compensating for processing delays. The forecasting loss is defined as $\mathcal{L}_{\text{forecast}}=\lambda_{\text{forecast}}\sum_{K}\sum_{t\in \{l,r\}}(|| G_{vf}^t - \hat{G}_{vf}^t ||_2^2 + || O_f^t - \hat{O}_f^t ||_2^2)$. In our implementation, we extend the calibration model by incorporating time embeddings to capture temporal dependencies.

\subsection{Training Losses}
\label{sec:traininglosses}

As illustrated in \cref{eq:multi_task_loss}, we adopt an end-to-end training strategy with a multi-task loss, introducing several modifications to the loss formulation to effectively accommodate the multi-task learning setting.

\begin{equation}
\begin{split}
  \mathcal{L} &= \mathcal{L}_{\text{origin}} 
               + \mathcal{L}_{\text{optical}}
               + \mathcal{L}_{\text{eye}} 
               + \mathcal{L}_{\text{seg}}
               + \mathcal{L}_{\text{depth}} \\
              &\quad + \mathcal{L}_{\text{visual}} 
               + \mathcal{L}_{\text{lookat}} 
               + \mathcal{L}_{\text{forecast}}
\end{split}
\label{eq:multi_task_loss}
\end{equation}

\noindent where the $\mathcal{L}_{\text{lookat}}$ is defined as

\begin{equation}
\begin{split}
  & \mathcal{L}_{\text{lookat}} = \lambda_{\text{lookat}} \, 
    \left\| P - \hat{P} \right\|_2^2, \\
  & \hat{P} = \frac{ (\hat{O}^l + s^l \hat{G}_v^l) 
                + (\hat{O}^r + s^r \hat{G}_v^r) }{2}
             \in \mathbb{R}^3, \\
  & s^t = \frac{P_z - \hat{O}_z^t}{\hat{G}_{vz}^t}, \quad t \in \{l,r\}.
\end{split}
\label{eq:loss_lookat}
\end{equation}

First, we compute the fixation point by intersecting the gaze vectors with the ground-truth depth, and supervise this fixation point. This design allows the predicted gaze of individual eyes to deviate from the true fixation point, consistent with physiological behavior, while ensuring that the fused fixation point remains accurate.

Second, for unsupervised depth, we leverage the available segmentation. Specifically, we apply these masks during loss computation so that the depth loss is calculated only within relevant local regions, such as the pupil or skin areas. This approach helps decouple loss contributions from different semantic components and improves performance.

 \section{Experiments}
\label{sec:experiments}

\subsection{Implementation Details}

By default, we employ ResNet50 \cite{he2016deep} as the backbone to extract features with a unet decoder. Furthermore, batch normalization (BN) layers in the backbone are replaced with group normalization (GN) \cite{wu2018group} to improve training stability. For the gaze decoder, used in both calibration and forecasting, we employ 6-layer transformer with 512-dimensional and 8-head self-attention. Positional encodings are incorporated into the input features to embed temporal information, which is essential for the forecasting task.  


We train the network using Adam optimizer \cite{kingma2014adam}($\beta_1$=0.9, $\beta_2$=0.999, weight decay=0.1) with a cosine learning rate scheduler (peak=0.001, warmup=1 epoch). Training runs for 30 epochs with batch size of 320 and input resolution of $160\times160$. Standard data augmentation (random cropping, color jittering, \etal) is applied to enhance generalization.



The loss weights are empirically determined as: $\lambda_{\text{origin}}$=1.0, $\lambda_{\text{optical}}$=2.0, $\lambda_{\text{eye}}$=1.0, $\lambda_{\text{visual}}$=5.0, $\lambda_{\text{focal}}$=5.0, $\lambda_{\text{dice}}$=1.0, $\lambda_{\text{cd}}$=0.5, $\lambda_{\text{ncc}}$=0.1, $\lambda_{\text{ssd}}$=0.1, $\lambda_{\text{ssim}}$=0.1, and $\lambda_{\text{lookat}}$=1.0. The gaze calibration module uses a 25\% masking ratio during training. In inference, we set $N$=40 for calibration or sequence samples, with no more than 20 calibration points, each represented by two frames.

\subsection{Gaze Estimation}

\begin{figure*}[t]
  \centering
   \includegraphics[width=1.0\linewidth]{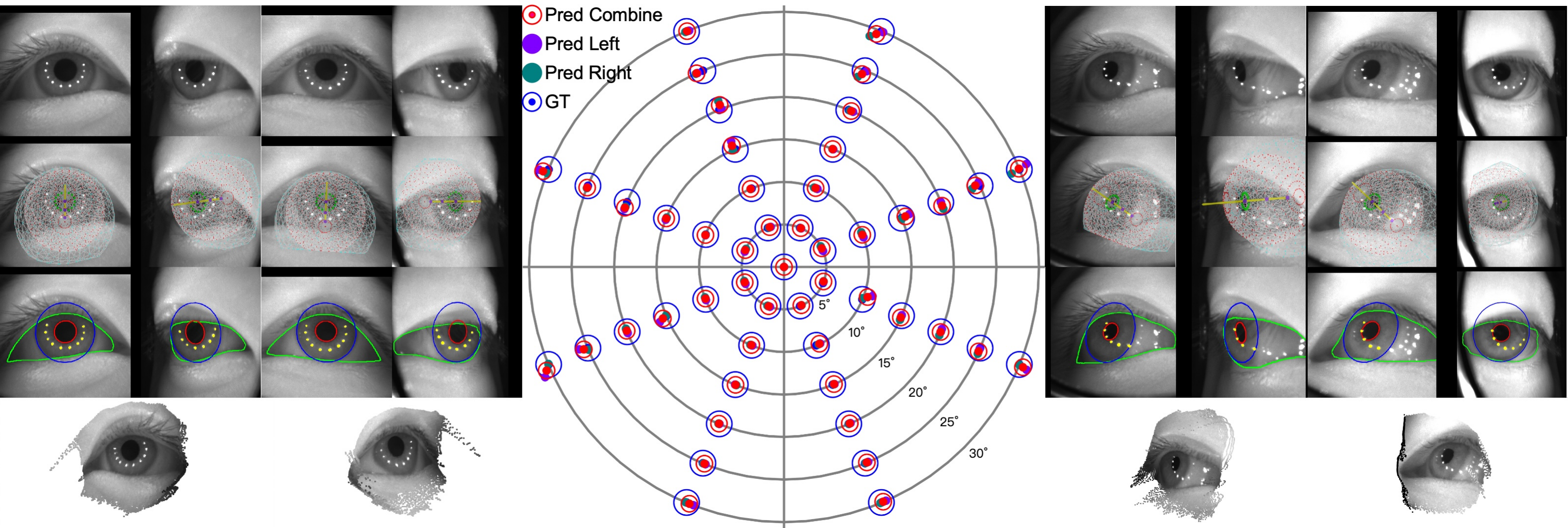}
   \caption{Qualitative results. The center shows the target point for the test subject, along with fixation points from the left, right, and their combination. Either side displays visualizations of additional model outputs and point cloud derived from the depth map.}
   \label{fig:qualitative_results}
\end{figure*}

The overall gaze performance is evaluated using accuracy, precision, origin distance, convergence error, and their corresponding P90 values, which indicate that most subjects have errors below this threshold. All test subjects are entirely independent from the training set. \cref{tab:gaze_results} reports the accuracy for both binocular and monocular settings, including combined gaze estimation as well as configurations using only the left or right eye. Performance is evaluated under three conditions: calibration-directly test, calibration-rewear after calibration, and without calibration. \textbf{The results demonstrate that our approach achieves state-of-the-art accuracy in both monocular and binocular settings, outperforming all previously reported gaze tracking methods to the best of our knowledge.} \cref{fig:qualitative_results} shows some qualitative results for the proposed method.

\begin{table}[htb]
\centering
\renewcommand{\arraystretch}{1.1}
\caption{\textbf{Gaze Forecasting.} Quantitative results of multi-step gaze prediction over 5 future time steps (90Hz). For each horizon, we report \emph{Accuracy/Origin} ($\si{\degree}$/mm).}
\label{tab:gaze_forecasting}
\begin{tabular*}{0.48\textwidth}{@{\extracolsep{\fill}}cccc}
Time Step & Combine & Left & Right \\
\specialrule{1.2pt}{0pt}{0pt}
1 & 0.48/0.25 & 0.53/0.28 & 0.53/0.26 \\
2 & 0.59/0.26 & 0.65/0.29 & 0.64/0.28 \\
3 & 0.60/0.26 & 0.65/0.29 & 0.64/0.28 \\
4 & 0.68/0.28 & 0.73/0.31 & 0.72/0.30 \\
5 & 0.65/0.25 & 0.70/0.31 & 0.70/0.30 \\
\end{tabular*}
\end{table}

\noindent \textbf{Gaze Forecasting.} As shown in \cref{tab:gaze_forecasting}, we perform gaze forecasting for the next five time steps. In application, to mitigate latency, the required gaze is estimated by interpolating between the predictions at these five timestamps.

\begin{table}[htb]
\centering
\renewcommand{\arraystretch}{1.1}
\caption{\textbf{Eye Parameters.} Euclidean Distance for Center/Radius error of Eyeball, Cornea and Pupil; Avg/P90 error for IPD(mm).}
\label{tab:gaze_parameters}
\begin{tabular*}{0.48\textwidth}{@{\extracolsep{\fill}}lcccc}
Camera & Eyeball & Cornea & Pupil & IPD \\
\specialrule{1.2pt}{0pt}{0pt}
Binocular & 0.35/0.22 & 0.19/0.12 & 0.10/0.02 & 0.07/0.11 \\
Monocular & 0.60/0.28 & 0.49/0.22 & 0.37/0.04 & 0.23/0.42 \\
\end{tabular*}
\end{table}

\noindent \textbf{Eye Parameters.} \cref{tab:gaze_parameters} presents the regression performance for eyeball parameters, as well as the inter pupillary distance (IPD) error, which is used for automatic IPD adjustment in head-mounted displays.
    
\begin{table}[htb]
\centering
\renewcommand{\arraystretch}{1.1}
\caption{\textbf{Eye Segmentation.} Mean IoU (mIoU) for four eye components (Sclera, Iris, Pupil, and Glint) for Binocular and Monocular.}
\label{tab:gaze_segmentation}
\begin{tabular*}{0.48\textwidth}{@{\extracolsep{\fill}}lcccc}
Camera & Sclera & Iris & Pupil & Glint\\
\specialrule{1.2pt}{0pt}{0pt}
Binocular & 0.9659 & 0.9570 & 0.9478 & 0.7899 \\
Monocular & 0.9708 & 0.9619 & 0.9467 & 0.7416 \\
\end{tabular*}
\end{table}

\noindent \textbf{Eye Segmentation.} \cref{tab:gaze_segmentation} reports the segmentation accuracy for eye regions, including sclera, cornea, pupil, and glints, evaluated using mIOU.

\subsection{Ablation Studies}

\begin{table}[htb]
\centering
\renewcommand{\arraystretch}{1.1}
\caption{\textbf{Ablation experiments for Transformer parameters.} 
$D$-number of layers, 
$W$-hidden dimension size, 
$H$-number of attention heads.}
\label{tab:model_ablation}
\begin{tabular*}{0.48\textwidth}{@{\extracolsep{\fill}}cc||cc||cc}
$D$ & Acc/Prec & $W$ & Acc/Prec & $H$ & Acc/Prec \\
\specialrule{1.2pt}{0pt}{0pt}
1 & 0.47/0.23 & 128 & 0.46/0.23 & 1 & 0.45/0.23 \\
2 & 0.45/0.23 & 256 & 0.45/0.23 & 2 & 0.45/0.23 \\
4 & 0.45/0.23 & \cellcolor{gray!20}\textbf{512} & \cellcolor{gray!20}\textbf{0.44/0.23} & 4 & 0.45/0.23 \\
\cellcolor{gray!20}\textbf{6} & \cellcolor{gray!20}\textbf{0.44/0.23} & 768 & 0.45/0.23 & \cellcolor{gray!20}\textbf{8} & \cellcolor{gray!20}\textbf{0.44/0.23} \\
8 & 0.44/0.23 & 1024 & 0.47/0.23 & 16 & 0.46/0.23 \\
\end{tabular*}
\end{table}

\noindent \textbf{Transformer Design.} Considering deployment on mobile device, we ablate our calibration model parameters. As shown in \cref{tab:model_ablation}, we compare different configurations of depth, width, and the number of attention heads to obtain the optimal parameters.

\begin{figure}[htb]
  \centering
  \includegraphics[width=\linewidth]{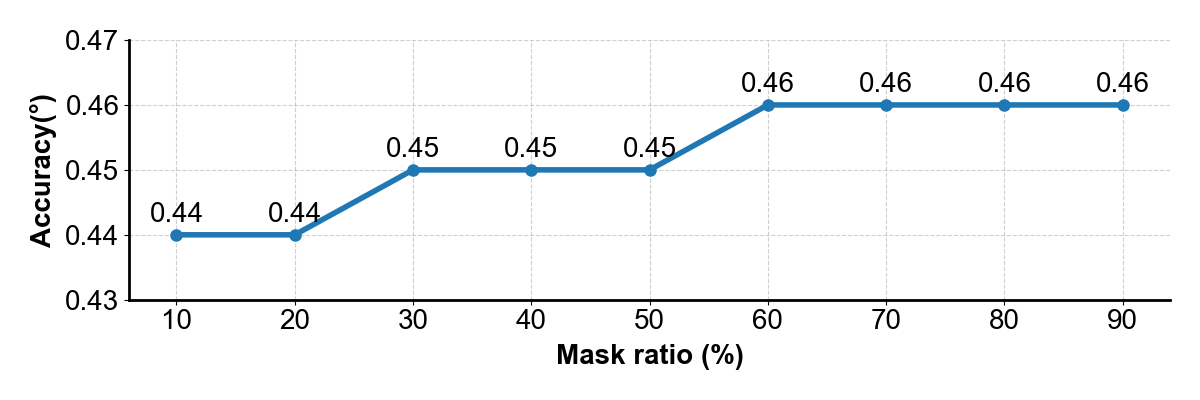}
  \caption{\textbf{Mask ratio.} For mask ratio, the accuracy of gaze calibration remains high, and the model exhibits great generalization.}
  \label{fig:accuracy_vs_mask_ratio}
\end{figure}

\noindent \textbf{Mask Ratio.} An important factor in the gaze calibration model is the mask ratio applied during training. As illustrated in \cref{fig:accuracy_vs_mask_ratio}, smaller ratio generally leads to higher inference accuracy. Nevertheless, even with larger mask ratio, the accuracy decreases by only 0.02°, highlighting the robustness and effectiveness of the proposed model design.

\begin{table}[htb]
\centering
\renewcommand{\arraystretch}{1.1}
\caption{\textbf{Ablation study of the camera parameters and ECA module,} illustrating its impact on the accuracy/origin, eyeball, cornea, and pupil estimations.}
\label{tab:camera}
\setlength{\tabcolsep}{2pt}
\begin{tabular*}{0.48\textwidth}{@{\extracolsep{\fill}}cccccc}
Cam & ECA & Gaze$\downarrow$ & Eyeball$\downarrow$ & Cornea$\downarrow$ & Pupil$\downarrow$ \\
\specialrule{1.2pt}{0pt}{0pt}
\ding{55} & \ding{55} & 0.53 / 0.25 & 0.88/0.29 & 0.81/0.22 & 0.73/0.05 \\
\checkmark & \ding{55} & 0.44 / 0.13 & 0.37/0.24 & 0.23/0.15 & 0.11/0.03 \\
\checkmark & \checkmark & \cellcolor{gray!20}\textbf{0.44 / 0.12} & \cellcolor{gray!20}\textbf{0.35/0.22} & \cellcolor{gray!20}\textbf{0.19/0.12} & \cellcolor{gray!20}\textbf{0.10/0.02} \\
\end{tabular*}
\end{table}

\noindent \textbf{Camera and ECA.} The epipolar constraint attention module is designed to fuse multi-view features, thereby enhancing the representation of feature. As shown in \cref{tab:camera}, removing the camera or ECA leads to a noticeable drop in the performance.

\section{Discussions}
\label{sec:discussions}


This work addresses a key challenge in gaze estimation by proposing a calibration model that integrates effectively into a unified, end-to-end framework. It supports accurate hand–eye interaction, eye-tracked foveated rendering, and auto-focus, enabling more natural and responsive immersive experiences in MR. The efficiency and precision of the proposed algorithm make it particularly suitable for mobile deployment, and it has already been successfully implemented in a commercial MR device.  


\noindent \textbf{Limitations.} The depth maps lack ground truth, preventing quantitative evaluation of reconstruction accuracy, and fine eye details (\eg, eyelashes and eyelids) are reconstructed with limited fidelity. The forecasting model remains less precise than the calibration stage. For the EyeID task, multi-task training significantly underperforms compared to a model trained solely for EyeID, indicating a potential direction for future work.

\noindent \textbf{Comparisons.} Reported accuracies on public near-eye datasets remain inadequate for MR interaction, likely due to limited label precision, insufficient coverage and diversity, weak standardization, and incomplete metadata. As a result, many methods still miss interaction-grade accuracy, for example, recent TEyeD methods report cross-validation errors above 2°. Since most prior datasets and methods target a single task, we do not make direct comparisons for our unified multi-task MR model. Instead, we provide high-precision ground truth ($<$0.15°) and release the acquisition procedure to support future benchmarking.

\section{Conclusions}
\label{sec:conclusions}

\ifreview
\else
We present \textit{PicoEyes}, a unified gaze estimation framework that can effectively processing both multicular and monocular inputs to estimate the full 3D structure of the eye, perform gaze calibration, and forecast gaze in an end-to-end manner. 
\fi
In addition, we introduce a large scale multi-view eye dataset that includes comprehensive 2D and 3D annotations related to eye tracking, as well as data for train, test, and calibration under various conditions. Experiments conducted on our proprietary datasets demonstrate that the proposed approach consistently outperforms previously reported gaze tracking methods, achieving state-of-the-art performance. 
\ifreview
\else
With the release of \textit{PicoEyes}, we aim to accelerate progress in gaze estimation and promote the development of methods that are both more accurate and more efficient.
\fi
\label{sec:picoeyes}

{
    \small
    \bibliographystyle{ieeenat_fullname}
    \bibliography{main}

@String(ECCV= {Eur. Conf. Comput. Vis.})

@String(ECCV  = {ECCV})

@article{guestrin2006general,
  title={General theory of remote gaze estimation using the pupil center and corneal reflections},
  author={Guestrin, Elias Daniel and Eizenman, Moshe},
  journal={IEEE Transactions on biomedical engineering},
  volume={53},
  number={6},
  pages={1124--1133},
  year={2006},
  publisher={IEEE}
}

@inproceedings{krafka2016eye,
  title={Eye tracking for everyone},
  author={Krafka, Kyle and Khosla, Aditya and Kellnhofer, Petr and Kannan, Harini and Bhandarkar, Suchendra and Matusik, Wojciech and Torralba, Antonio},
  booktitle={Proceedings of the IEEE conference on computer vision and pattern recognition},
  pages={2176--2184},
  year={2016}
}

@inproceedings{bao2025gazegene,
  title={GazeGene: Large-scale Synthetic Gaze Dataset with 3D Eyeball Annotations},
  author={Bao, Yiwei and Wang, Zhiming and Lu, Feng},
  booktitle={Proceedings of the Computer Vision and Pattern Recognition Conference},
  pages={18749--18759},
  year={2025}
}

@inproceedings{zhang2015appearance,
  title={Appearance-based gaze estimation in the wild},
  author={Zhang, Xucong and Sugano, Yusuke and Fritz, Mario and Bulling, Andreas},
  booktitle={Proceedings of the IEEE conference on computer vision and pattern recognition},
  pages={4511--4520},
  year={2015}
}

@article{zhang2017mpiigaze,
  title={Mpiigaze: Real-world dataset and deep appearance-based gaze estimation},
  author={Zhang, Xucong and Sugano, Yusuke and Fritz, Mario and Bulling, Andreas},
  journal={IEEE transactions on pattern analysis and machine intelligence},
  volume={41},
  number={1},
  pages={162--175},
  year={2017},
  publisher={IEEE}
}

@inproceedings{guo2019generalized,
  title={A generalized and robust method towards practical gaze estimation on smart phone},
  author={Guo, Tianchu and Liu, Yongchao and Zhang, Hui and Liu, Xiabing and Kwak, Youngjun and In Yoo, Byung and Han, Jae-Joon and Choi, Changkyu},
  booktitle={Proceedings of the IEEE/CVF International Conference on Computer Vision Workshops},
  pages={0--0},
  year={2019}
}

@inproceedings{zhang2020eth,
  title={Eth-xgaze: A large scale dataset for gaze estimation under extreme head pose and gaze variation},
  author={Zhang, Xucong and Park, Seonwook and Beeler, Thabo and Bradley, Derek and Tang, Siyu and Hilliges, Otmar},
  booktitle={European conference on computer vision},
  pages={365--381},
  year={2020},
  organization={Springer}
}

@inproceedings{hisadome2024rotation,
  title={Rotation-constrained cross-view feature fusion for multi-view appearance-based gaze estimation},
  author={Hisadome, Yoichiro and Wu, Tianyi and Qin, Jiawei and Sugano, Yusuke},
  booktitle={Proceedings of the IEEE/CVF Winter Conference on Applications of Computer Vision},
  pages={5985--5994},
  year={2024}
}

@inproceedings{jindal2024spatio,
  title={Spatio-temporal attention and gaussian processes for personalized video gaze estimation},
  author={Jindal, Swati and Yadav, Mohit and Manduchi, Roberto},
  booktitle={Proceedings of the IEEE/CVF Conference on Computer Vision and Pattern Recognition},
  pages={604--614},
  year={2024}
}

@article{xu2024camco,
  title={Camco: Camera-controllable 3d-consistent image-to-video generation},
  author={Xu, Dejia and Nie, Weili and Liu, Chao and Liu, Sifei and Kautz, Jan and Wang, Zhangyang and Vahdat, Arash},
  journal={arXiv preprint arXiv:2406.02509},
  year={2024}
}

@inproceedings{he2025cameractrl,
  title={Cameractrl: Enabling camera control for video diffusion models},
  author={He, Hao and Xu, Yinghao and Guo, Yuwei and Wetzstein, Gordon and Dai, Bo and Li, Hongsheng and Yang, Ceyuan},
  booktitle={The Thirteenth International Conference on Learning Representations},
  year={2025}
}

@inproceedings{schops2020having,
  title={Why having 10,000 parameters in your camera model is better than twelve},
  author={Schops, Thomas and Larsson, Viktor and Pollefeys, Marc and Sattler, Torsten},
  booktitle={Proceedings of the IEEE/CVF Conference on Computer Vision and Pattern Recognition},
  pages={2535--2544},
  year={2020}
}

@article{jia2020plucker,
  title={Pl{\"u}cker coordinates for lines in the space},
  author={Jia, Yan-Bin},
  journal={Problem Solver Techniques for Applied Computer Science, Com-S-477/577 Course Handout},
  volume={3},
  year={2020},
  publisher={Iowa State University Ames, IA, USA}
}

@inproceedings{yao2018mvsnet,
  title={Mvsnet: Depth inference for unstructured multi-view stereo},
  author={Yao, Yao and Luo, Zixin and Li, Shiwei and Fang, Tian and Quan, Long},
  booktitle={Proceedings of the European conference on computer vision (ECCV)},
  pages={767--783},
  year={2018}
}

@article{guo2024stereo,
  title={Stereo anything: Unifying stereo matching with large-scale mixed data},
  author={Guo, Xianda and Zhang, Chenming and Zhang, Youmin and Nie, Dujun and Wang, Ruilin and Zheng, Wenzhao and Poggi, Matteo and Chen, Long},
  journal={arXiv preprint arXiv:2411.14053},
  year={2024}
}

@inproceedings{ronneberger2015u,
  title={U-net: Convolutional networks for biomedical image segmentation},
  author={Ronneberger, Olaf and Fischer, Philipp and Brox, Thomas},
  booktitle={International Conference on Medical image computing and computer-assisted intervention},
  pages={234--241},
  year={2015},
  organization={Springer}
}

@inproceedings{lin2017focal,
  title={Focal loss for dense object detection},
  author={Lin, Tsung-Yi and Goyal, Priya and Girshick, Ross and He, Kaiming and Doll{\'a}r, Piotr},
  booktitle={Proceedings of the IEEE international conference on computer vision},
  pages={2980--2988},
  year={2017}
}

@inproceedings{milletari2016v,
  title={V-net: Fully convolutional neural networks for volumetric medical image segmentation},
  author={Milletari, Fausto and Navab, Nassir and Ahmadi, Seyed-Ahmad},
  booktitle={2016 fourth international conference on 3D vision (3DV)},
  pages={565--571},
  year={2016},
  organization={Ieee}
}

@inproceedings{fan2017point,
  title={A point set generation network for 3d object reconstruction from a single image},
  author={Fan, Haoqiang and Su, Hao and Guibas, Leonidas J},
  booktitle={Proceedings of the IEEE conference on computer vision and pattern recognition},
  pages={605--613},
  year={2017}
}

@inproceedings{lewis1995fast,
  title={Fast template matching},
  author={Lewis, John P and others},
  booktitle={Vision interface},
  volume={95},
  number={120123},
  pages={15--19},
  year={1995},
  organization={Quebec City, QC, Canada}
}

@article{horn1981determining,
  title={Determining optical flow},
  author={Horn, Berthold KP and Schunck, Brian G},
  journal={Artificial intelligence},
  volume={17},
  number={1-3},
  pages={185--203},
  year={1981},
  publisher={Elsevier}
}

@article{wang2004image,
  title={Image quality assessment: from error visibility to structural similarity},
  author={Wang, Zhou and Bovik, Alan C and Sheikh, Hamid R and Simoncelli, Eero P},
  journal={IEEE transactions on image processing},
  volume={13},
  number={4},
  pages={600--612},
  year={2004},
  publisher={IEEE}
}

@inproceedings{he2022masked,
  title={Masked autoencoders are scalable vision learners},
  author={He, Kaiming and Chen, Xinlei and Xie, Saining and Li, Yanghao and Doll{\'a}r, Piotr and Girshick, Ross},
  booktitle={Proceedings of the IEEE/CVF conference on computer vision and pattern recognition},
  pages={16000--16009},
  year={2022}
}

@inproceedings{devlin2019bert,
  title={Bert: Pre-training of deep bidirectional transformers for language understanding},
  author={Devlin, Jacob and Chang, Ming-Wei and Lee, Kenton and Toutanova, Kristina},
  booktitle={Proceedings of the 2019 conference of the North American chapter of the association for computational linguistics: human language technologies, volume 1 (long and short papers)},
  pages={4171--4186},
  year={2019}
}

@article{radford2018improving,
  title={Improving language understanding by generative pre-training},
  author={Radford, Alec and Narasimhan, Karthik and Salimans, Tim and Sutskever, Ilya and others},
  year={2018},
  publisher={San Francisco, CA, USA}
}

@inproceedings{he2016deep,
  title={Deep residual learning for image recognition},
  author={He, Kaiming and Zhang, Xiangyu and Ren, Shaoqing and Sun, Jian},
  booktitle={Proceedings of the IEEE conference on computer vision and pattern recognition},
  pages={770--778},
  year={2016}
}

@inproceedings{wu2018group,
  title={Group normalization},
  author={Wu, Yuxin and He, Kaiming},
  booktitle={Proceedings of the European conference on computer vision (ECCV)},
  pages={3--19},
  year={2018}
}

@article{kingma2014adam,
  title={Adam: A method for stochastic optimization},
  author={Kingma, Diederik P},
  journal={arXiv preprint arXiv:1412.6980},
  year={2014}
}

@article{ravi2024sam,
  title={Sam 2: Segment anything in images and videos},
  author={Ravi, Nikhila and Gabeur, Valentin and Hu, Yuan-Ting and Hu, Ronghang and Ryali, Chaitanya and Ma, Tengyu and Khedr, Haitham and R{\"a}dle, Roman and Rolland, Chloe and Gustafson, Laura and others},
  journal={arXiv preprint arXiv:2408.00714},
  year={2024}
}

@inproceedings{rekimoto2025gazellm,
  title={GazeLLM: Multimodal LLMs incorporating human visual attention},
  author={Rekimoto, Jun},
  booktitle={Proceedings of the Augmented Humans International Conference 2025},
  pages={302--311},
  year={2025}
}

@article{yan2024voila,
  title={Voila-a: Aligning vision-language models with user's gaze attention},
  author={Yan, Kun and Wang, Zeyu and Ji, Lei and Wang, Yuntao and Duan, Nan and Ma, Shuai},
  journal={Advances in Neural Information Processing Systems},
  volume={37},
  pages={1890--1918},
  year={2024}
}

@article{liu2025fovealnet,
  title={Fovealnet: Advancing ai-driven gaze tracking solutions for efficient foveated rendering in virtual reality},
  author={Liu, Wenxuan and Duinkharjav, Budmonde and Sun, Qi and Zhang, Sai Qian},
  journal={IEEE Transactions on Visualization and Computer Graphics},
  year={2025},
  publisher={IEEE}
}

@inproceedings{pfeuffer2017gaze+,
  title={Gaze+ pinch interaction in virtual reality},
  author={Pfeuffer, Ken and Mayer, Benedikt and Mardanbegi, Diako and Gellersen, Hans},
  booktitle={Proceedings of the 5th symposium on spatial user interaction},
  pages={99--108},
  year={2017}
}

@article{chakravarthula2018focusar,
  title={Focusar: Auto-focus augmented reality eyeglasses for both real world and virtual imagery},
  author={Chakravarthula, Praneeth and Dunn, David and Ak{\c{s}}it, Kaan and Fuchs, Henry},
  journal={IEEE transactions on visualization and computer graphics},
  volume={24},
  number={11},
  pages={2906--2916},
  year={2018},
  publisher={IEEE}
}

@article{palmero2020openeds2020,
  title={Openeds2020: Open eyes dataset},
  author={Palmero, Cristina and Sharma, Abhishek and Behrendt, Karsten and Krishnakumar, Kapil and Komogortsev, Oleg V and Talathi, Sachin S},
  journal={arXiv preprint arXiv:2005.03876},
  year={2020}
}

@inproceedings{kim2019nvgaze,
  title={Nvgaze: An anatomically-informed dataset for low-latency, near-eye gaze estimation},
  author={Kim, Joohwan and Stengel, Michael and Majercik, Alexander and De Mello, Shalini and Dunn, David and Laine, Samuli and McGuire, Morgan and Luebke, David},
  booktitle={Proceedings of the 2019 CHI conference on human factors in computing systems},
  pages={1--12},
  year={2019}
}

@inproceedings{kellnhofer2019gaze360,
  title={Gaze360: Physically unconstrained gaze estimation in the wild},
  author={Kellnhofer, Petr and Recasens, Adria and Stent, Simon and Matusik, Wojciech and Torralba, Antonio},
  booktitle={Proceedings of the IEEE/CVF international conference on computer vision},
  pages={6912--6921},
  year={2019}
}

@inproceedings{zhang2017s,
  title={It's written all over your face: Full-face appearance-based gaze estimation},
  author={Zhang, Xucong and Sugano, Yusuke and Fritz, Mario and Bulling, Andreas},
  booktitle={Proceedings of the IEEE conference on computer vision and pattern recognition workshops},
  pages={51--60},
  year={2017}
}

@article{wang2024pveye,
  title={PVEye: A Large Posture-Variant Eye Tracking Dataset for Head-Mounted AR Devices},
  author={Wang, Xiaodong and Bai, Xiaowei and Xie, Liang and Li, Yingxi and Wang, Qining and Yan, Ye and Yin, Erwei},
  journal={IEEE Transactions on Visualization and Computer Graphics},
  year={2024},
  publisher={IEEE}
}

@inproceedings{fuhl2021teyed,
  title={Teyed: Over 20 million real-world eye images with pupil, eyelid, and iris 2d and 3d segmentations, 2d and 3d landmarks, 3d eyeball, gaze vector, and eye movement types},
  author={Fuhl, Wolfgang and Kasneci, Gjergji and Kasneci, Enkelejda},
  booktitle={2021 IEEE International Symposium on Mixed and Augmented Reality (ISMAR)},
  pages={367--375},
  year={2021},
  organization={IEEE}
}

@article{wu2020magiceyes,
  title={Magiceyes: A large scale eye gaze estimation dataset for mixed reality},
  author={Wu, Zhengyang and Rajendran, Srivignesh and van As, Tarrence and Zimmermann, Joelle and Badrinarayanan, Vijay and Rabinovich, Andrew},
  journal={arXiv preprint arXiv:2003.08806},
  year={2020}
}

@inproceedings{cheng2023dvgaze,
  title={Dvgaze: Dual-view gaze estimation},
  author={Cheng, Yihua and Lu, Feng},
  booktitle={Proceedings of the IEEE/CVF International Conference on Computer Vision},
  pages={20632--20641},
  year={2023}
}

@article{garbin2019openeds,
  title={Openeds: Open eye dataset},
  author={Garbin, Stephan J and Shen, Yiru and Schuetz, Immo and Cavin, Robert and Hughes, Gregory and Talathi, Sachin S},
  journal={arXiv preprint arXiv:1905.03702},
  year={2019}
}

@inproceedings{wu2019eyenet,
  title={Eyenet: A multi-task deep network for off-axis eye gaze estimation},
  author={Wu, Zhengyang and Rajendran, Srivignesh and Van As, Tarrence and Badrinarayanan, Vijay and Rabinovich, Andrew},
  booktitle={2019 IEEE/CVF International Conference on Computer Vision Workshop (ICCVW)},
  pages={3683--3687},
  year={2019},
  organization={IEEE}
}

@article{hansen2009eye,
  title={In the eye of the beholder: A survey of models for eyes and gaze},
  author={Hansen, Dan Witzner and Ji, Qiang},
  journal={IEEE transactions on pattern analysis and machine intelligence},
  volume={32},
  number={3},
  pages={478--500},
  year={2009},
  publisher={IEEE}
}

@inproceedings{wood20163d,
  title={A 3d morphable eye region model for gaze estimation},
  author={Wood, Erroll and Baltru{\v{s}}aitis, Tadas and Morency, Louis-Philippe and Robinson, Peter and Bulling, Andreas},
  booktitle={European conference on computer vision},
  pages={297--313},
  year={2016},
  organization={Springer}
}

@article{cheng2024appearance,
  title={Appearance-based gaze estimation with deep learning: A review and benchmark},
  author={Cheng, Yihua and Wang, Haofei and Bao, Yiwei and Lu, Feng},
  journal={IEEE Transactions on Pattern Analysis and Machine Intelligence},
  volume={46},
  number={12},
  pages={7509--7528},
  year={2024},
  publisher={IEEE}
}

@inproceedings{linden2019learning,
  title={Learning to personalize in appearance-based gaze tracking},
  author={Lind{\'e}n, Erik and Sjostrand, Jonas and Proutiere, Alexandre},
  booktitle={Proceedings of the IEEE/CVF international conference on computer vision workshops},
  pages={0--0},
  year={2019}
}
}

\clearpage
\setcounter{page}{1}
\maketitlesupplementary


\section*{Supplementary Analysis of Data Distribution}

\begin{figure}[htp]
    \centering
    \includegraphics[width=0.77\linewidth]{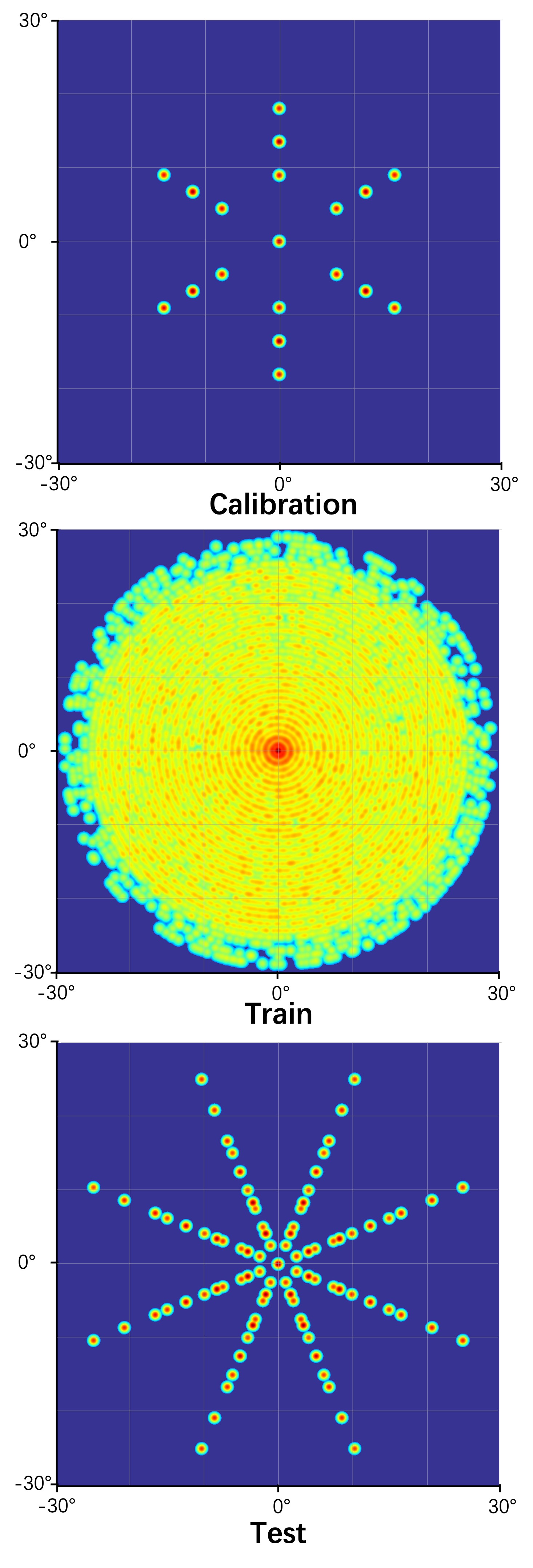}
    \caption{The gaze angle distribution of our dataset.}
    \label{fig:data_distribution}
\end{figure}

In \cref{fig:data_distribution}, we illustrate the gaze angle distributions for the calibration, train, and test subsets of our dataset, shown from top to bottom. As depicted, gaze samples are uniformly and comprehensively distributed across the angular range of [-30$\si{\degree}$, 30$\si{\degree}$] along both the horizontal (X) and vertical (Y) axes. This balanced coverage ensures that all datasets encompass a wide variety of gaze directions, which is beneficial for enhancing model robustness and generalization.

\begin{figure*}[htp]
    \centering
    \includegraphics[width=0.9\linewidth]{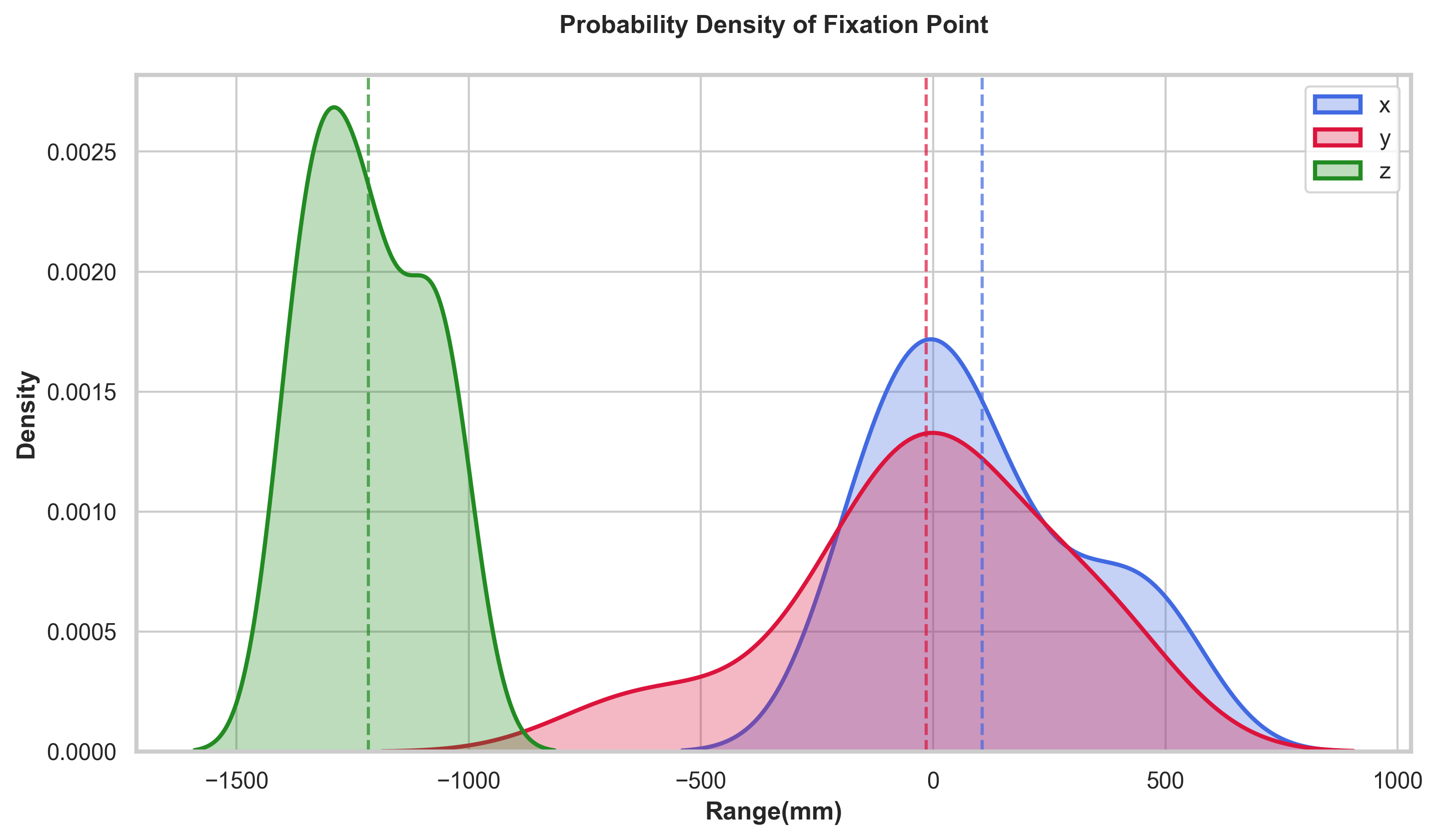}
    \caption{Density map of the fixation point in our dataset.}
    \label{fig:lookat_kde}
\end{figure*}

\begin{figure*}[htp]
    \centering
    \includegraphics[width=0.9\linewidth]{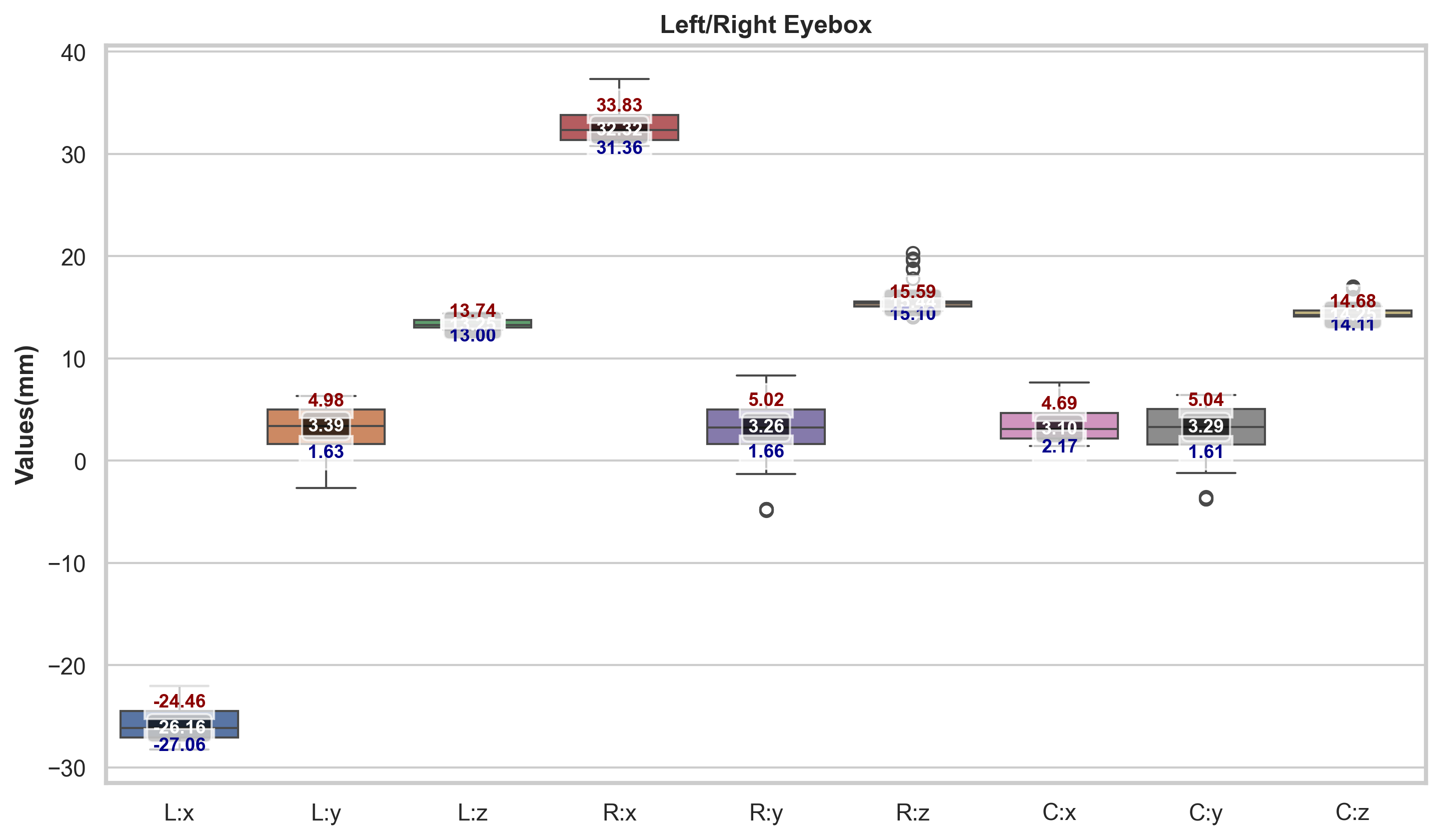}
    \caption{The eyebox distribution of our dataset.}
    \label{fig:eyebox_distribution}
\end{figure*}

\begin{figure*}[htp]
    \centering
    \includegraphics[width=0.9\linewidth]{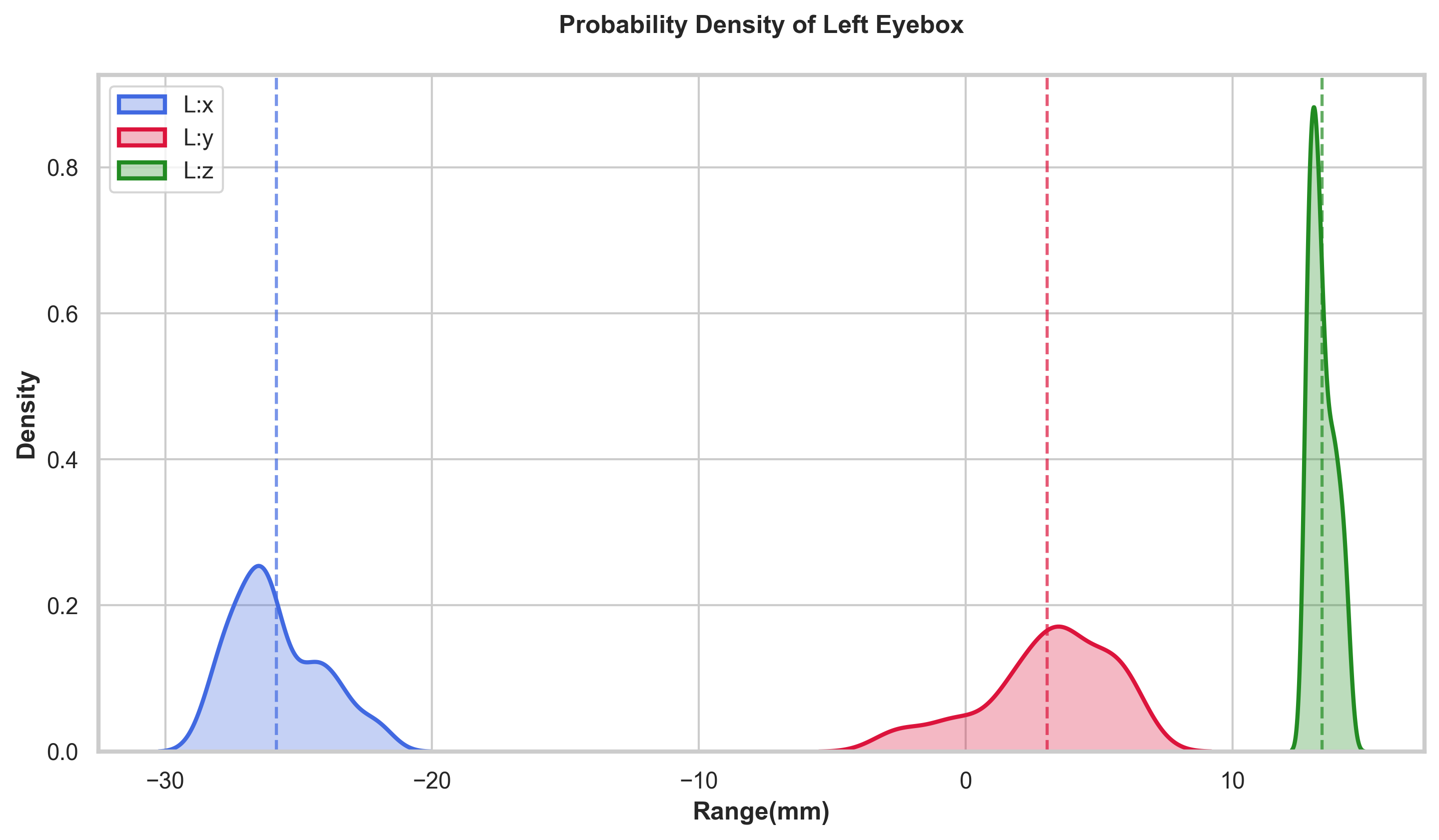}
    \caption{Density map of the left eyebox in our dataset.}
    \label{fig:left_eyebox}
\end{figure*}

\begin{figure*}[htp]
    \centering
    \includegraphics[width=0.9\linewidth]{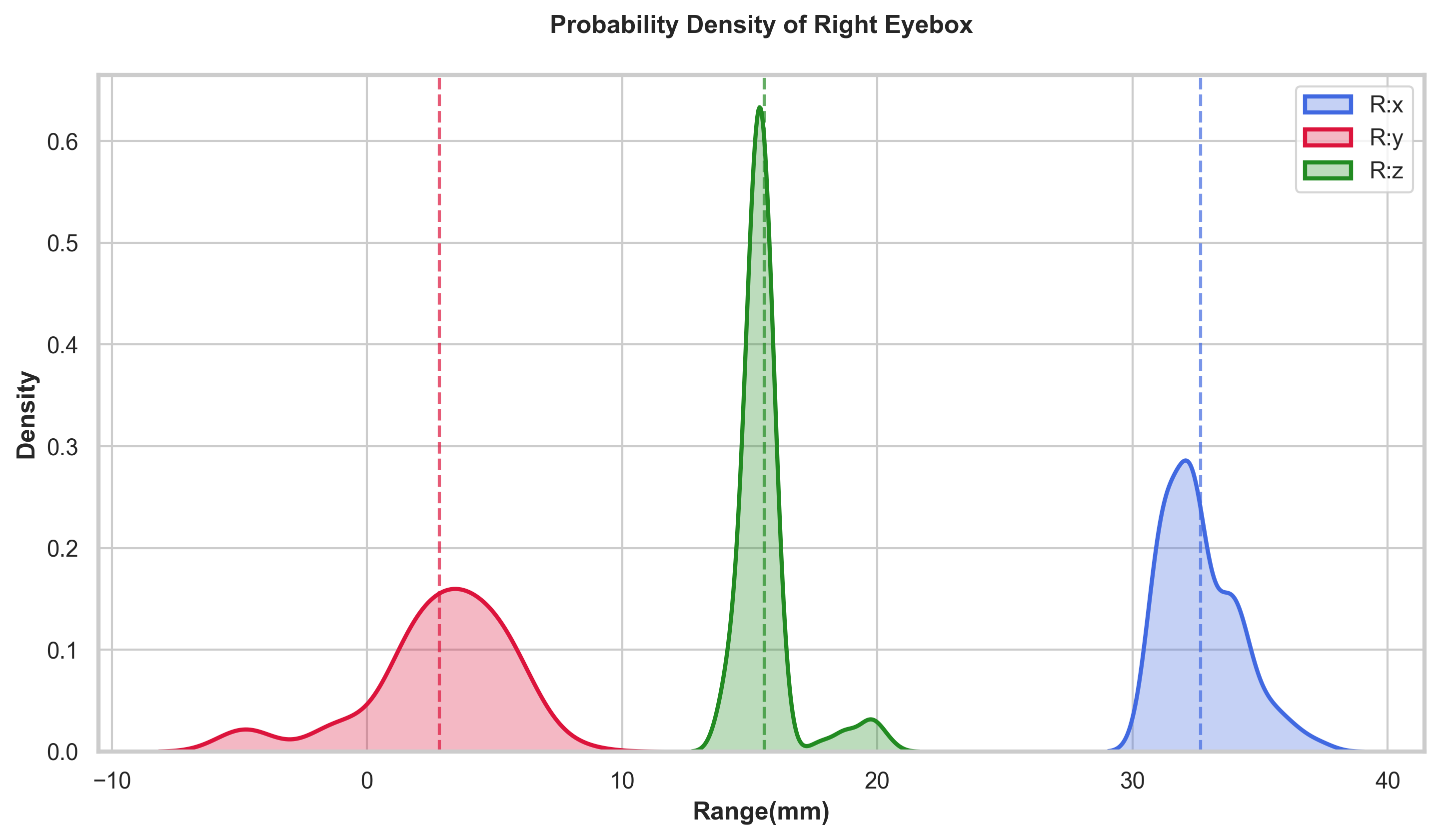}
    \caption{Density map of the right eyebox in our dataset.}
    \label{fig:right_eyebox}
\end{figure*}

The 3D spatial distribution of fixation points is visualized in \cref{fig:lookat_kde}. The overall spatial coverage of the eyebox is illustrated in \cref{fig:eyebox_distribution}.  For clarity, separate density maps of the left and right eyeboxes are shown in \cref{fig:left_eyebox} and \cref{fig:right_eyebox}, respectively.

\section*{Additional Dataset Visualizations}

\begin{figure*}
    \centering
    \includegraphics[width=0.85\linewidth]{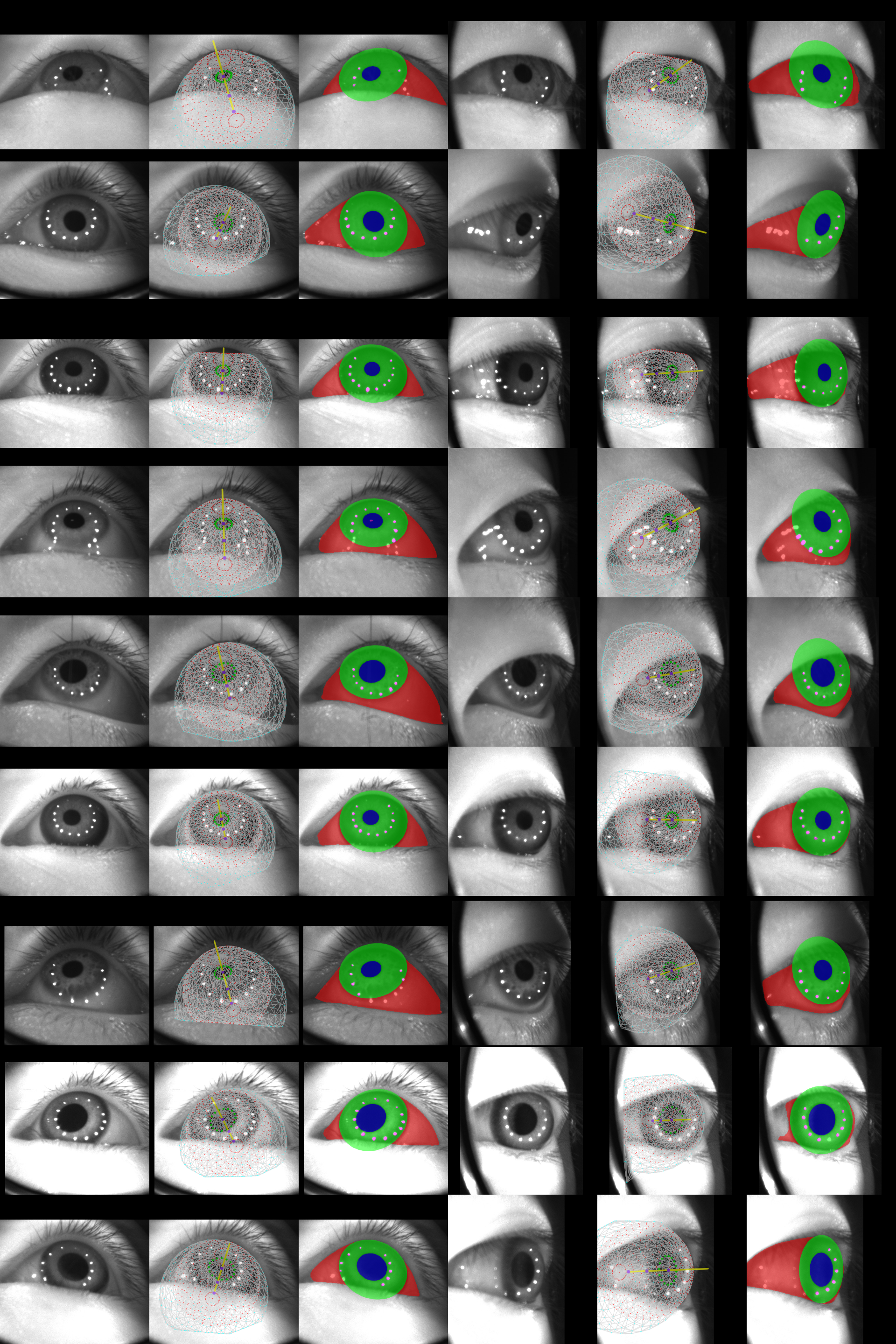}
    \caption{Example images and annotations from our dataset.}
    \label{fig:data_examples}
\end{figure*}

In \cref{fig:data_examples}, we present qualitative visualizations that include both eye segmentation maps and 3D structures. The segmentation maps delineate precise pixel‑level components of the eye (sclera, iris, pupil, and glints), while the 3D models faithfully reconstruct the geometric structure. or comparison, the left three columns and the right three columns display eye images captured simultaneously by two distinct cameras, highlighting variations in viewpoint and appearance across sensors.

\section*{Privacy Protection and Public Release Plan}

To mitigate potential privacy concerns, all eye images in this paper have been preprocessed and are not presented in their raw form. Regarding the public release of data and code, both may contain information that could reveal participants’ biometric characteristics, therefore, they will only be made available after thorough de-identification to remove all personally identifiable elements. In addition, because the codebase includes numerous optimizations tailored for deployment on mobile platforms, such as model quantization and pruning, these components will be removed prior to release. The final open source version will retain all elements necessary to reproduce the results reported in this study, thereby ensuring both reproducibility and the protection of participant privacy.

\end{document}